\documentclass[draft]{agujournal2019}
\usepackage{url} 
\usepackage{lineno}
\usepackage[inline]{trackchanges} 
\usepackage{soul}
\draftfalse
\justifying
\usepackage{rotating}
\usepackage{algorithm2e}
\usepackage{verbatim}
\usepackage{lscape}
\usepackage{graphics}
\usepackage{amsmath}
\usepackage{lscape}
\usepackage{supertabular}
\usepackage{hhline}
\usepackage{multirow}
\usepackage{mathtools,amssymb}
\usepackage{algpseudocode}

\journalname{}

\begin{document}

\title{At the junction between deep learning and statistics of extremes: formalizing the landslide hazard definition}

\authors{Ashok Dahal\affil{1},
Rapha\"{e}l Huser\affil{2},
 Luigi Lombardo\affil{1}}

 \affiliation{1}{University of Twente, Faculty of Geo-Information Science and Earth Observation (ITC), PO Box 217, Enschede, AE 7500, Netherlands}
 \affiliation{2}{Statistics Program, Computer, Electrical and Mathematical Sciences and Engineering (CEMSE) Division,\\ King Abdullah University of Science and Technology (KAUST), Thuwal 23955-6900, Saudi Arabia}

\correspondingauthor{Ashok Dahal}{a.dahal@utwente.nl}

\begin{keypoints}
\item A Landslide hazard model is developed by combining deep learning and the extended generalized Pareto distribution from extreme value theory.
\item Thirty years of past observations are used to understand landslide hazard in Nepal.
\item Landslide hazard is predicted for multiple return periods based on climate projections up to the end of the century.
\end{keypoints}

\begin{abstract}
The most adopted definition of landslide hazard combines spatial information about landslide location (susceptibility), threat (intensity), and frequency (return period). Only the first two elements are usually considered and estimated when working over vast areas. Even then, separate models constitute the standard, with frequency being rarely investigated. Frequency and intensity are intertwined and depend on each other because larger events occur less frequently and vice versa. However, due to the lack of multi-temporal inventories and joint statistical models, modelling such properties via a unified hazard model has always been challenging and has yet to be attempted. Here, we develop a unified model to estimate landslide hazard at the slope unit level to address such gaps. We employed deep learning, combined with a model motivated by extreme-value theory to analyse an inventory of 30 years of observed rainfall-triggered landslides in Nepal and assess landslide hazard for multiple return periods. We also use our model to further explore landslide hazard for the same return periods under different climate change scenarios up to the end of the century. Our results show that the proposed model performs excellently and can be used to model landslide hazard in a unified manner. Geomorphologically, we find that under both climate change scenarios (SSP245 and SSP885), landslide hazard is likely to increase up to two times on average in the lower Himalayan regions while remaining the same in the middle Himalayan region whilst decreasing slightly in the upper Himalayan region areas.
\end{abstract}


%
%
%
%
%
%

%

\section{Introduction}
\label{sec:Introduction}
Landslides are natural processes that contribute to the evolution of any given landscape. However, as our society has increasingly sprawled over time towards mountainous terrains, their occurrence threatens lives and infrastructures \cite{petley2012global,ozturk2022climate}. 
A common approach to mitigate their adverse effects is to model their failure and, on this basis, either plan appropriate mitigation actions or assign a different land use. The term ``model", though, is a broad umbrella encompassing different approaches.
To reconstruct the complete landslide phenomenon, physically-based models are usually the appropriate tools, as they can simulate landslide initiation, propagation, and runout \cite{mergili2020back,pudasaini2021mechanics}. However, these tools are also highly data-demanding 
because the initial conditions of the relevant hydro-mechanical equations, such as geotechnical parameters and triggering factors (e.g., heavy precipitation intensity or earthquake shaking levels) vary in space and time and must be precisely inferred from data. 
This is why the application of physically-based models has been confined to small areas, up to single catchments and rarely beyond this level \cite{Bout2021a}. Contrastingly, 
statistical models have been used to estimate landslide-related information over broader areas. However, the prediction target has historically differed from what physically-based models have focused on. Specifically, out of the most accepted landslide definition---where, when or how frequently landslides may occur, along with their expected intensity \cite{guzzetti1999landslidehazard}---
statistical models have almost exclusively addressed the ``where'' component. This component is traditionally referred to as the landslide susceptibility \cite <see,>[]{reichenbach-2018}. As for the ``when'' or ``how frequently'' components, the scientific efforts have consolidated into near-real-time or early-warning systems such as the Ground Failure module of the USGS \cite{nowicki2018global} for co-seismic landslides or the LHASA product of NASA \cite{stanley2021data} for rainfall-induced landslides. Finally, regarding the intensity component \cite{corominas2014recommendations}, this is the element that has historically hardly received any attention from the 
statistical perspective, with limited models tested for rockfalls \cite{lari2014} and debris flows \cite{lombardo2018modeling,Lombardo.etal:2018,lombardo2019numerical}. 

Irrespective of the number and quality of the research context mentioned above, hardly anybody has jointly modelled the three components altogether or even modelled those three components within the same general framework except for the seminal work of \citeA{guzzetti2005probabilistic}. Recently, \citeA{aguilera2022prediction} and \citeA{bryce2022unified} linked the spatial component of the hazard definition to the expected size of a landslide. Furthermore, the landslide size, which is a proxy for the landslide intensity, has been jointly modeled together with landslide counts or density per unit area by \citeA{yadav2022joint}, but this approach still does not fully comply with the landslide hazard definition. No database of other metrics such as velocity, kinetic energy and impact force exists for landslides distributed over large areas. Despite this methodological progress, the temporal dimension was rarely accounted for. This aspect has been recently tackled by \citeA{lombardo2020space}, \citeA{dahal2022space}, and \citeA{fang2024landslide}. In particular, 
\citeA{dahal2022space} adopted a machine learning perspective and implemented an Ensemble Neural Network (ENN) in their work. The ensemble performed susceptibility and landslide area density estimation jointly over four years following the Gorkha earthquake in Nepal. Their architecture relied on a regular partition of the Nepali landscape, following the same $1 \times 1$ km$^2$ data structure of the multi-temporal landslide database mapped by \citeA{kincey2021evolution}. However, in a 1~km$^2$, a landscape can host both streamlines and ridges, making such a regular subdivision far from being representative of the actual landscape morphology. Moreover, due to the nature of the statistical distribution in consideration as well as the limited temporal observation period, their model had limited capability in estimating the tail properties and in extrapolating in time.

In this work, we develop a novel formal framework to solve the landslide hazard modelling problem on a regional scale by combining a model motivated by extreme-value theory with deep learning, which facilitates the extraction of complex information from data. Specifically, to model the landslide intensity, we use the extended Generalised Pareto Distribution (eGPD; \citeA{papastathopoulos2013extended,naveau2016modeling}), which respects the heavy-tailed nature of the landslide size (here, represented by the area density) distribution and has an appropriate tail behavior both in the lower and the upper tail, with a smooth transition in between. Two main reasons for using such a statistical model in our context are that (i) it naturally handles heavy-tailed data, while describing the full data range (lower tail, bulk, upper tail) in a parsimonious way; and (ii) it can be used to extrapolate further into the tail to estimate potentially high landslide intensities using a mathematically-justified framework. Our work also extends the analyses over the 30-year time window that  \citeA{jones202130} have used to map landslides yearly, enabling one to ``see'' much further into the past and, therefore, provide more reliable estimates. With the fitted model, we further analyse the landslide hazard under two climate change scenarios: the middle pathway (SSP245) and fossil-fueled development (SSP585) scenarios.

\section{Study Area and Inventory}
\label{sec:studyarea}

Nepal is a mountainous country overlooked by part of the Himalayan belt. 
Its elevation ranges from 60 m to 8848 m within an average span of 130 km. 
As a result, almost 82\% of the country is classified as mountainous or hilly with gentle to extreme slopes, and its territory is frequently affected by landslides due to earthquakes and extreme precipitation.
For this reason, slope instabilities in Nepal have been the focus of numerous research studies, also stressing other relevant predisposing and causative factors such as unplanned construction \cite{tiwari2014influence}, road cuts \cite{mcadoo2018roads}, and irrigation \cite{gerrard2002relationships}. 

\citeA{jones202130} produced a multitemporal landslide inventory for a large portion of Nepal, offering a unique possibility to examine the spatiotemporal distribution of slope failures and their characteristics. Therefore, to model the spatiotemporal landslide hazard in this study, we used the mapping extent of the \citeA{jones202130} inventory as our study area. 

\begin{figure}[t!]
        \centering
	\includegraphics[scale=1.0]{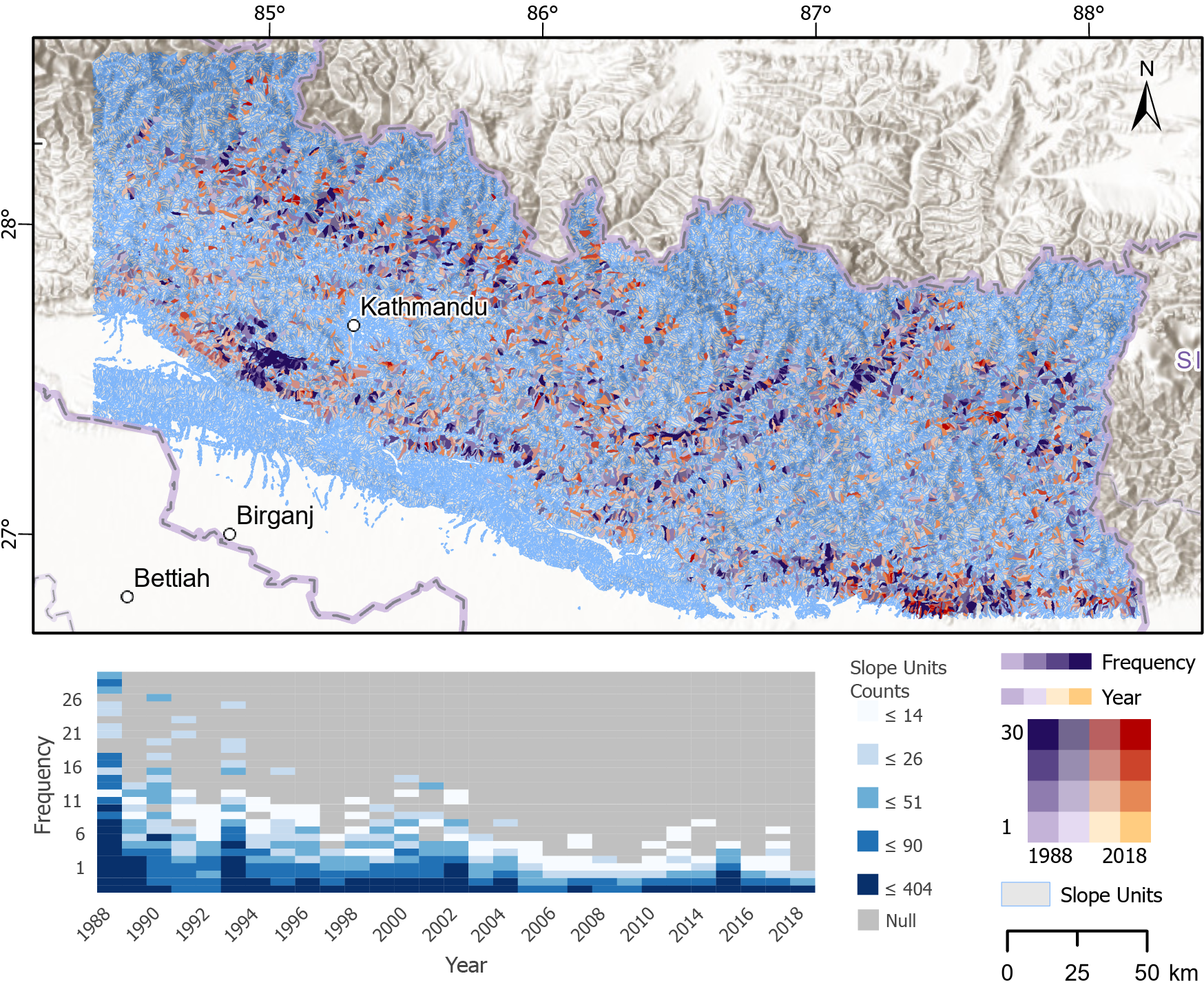}
	\caption{Map~showing the study area, mapping units, and observed landslide inventory  (top) for 1988--2018. The bivariate histogram plot (bottom right) shows the intensity of landslides in individual slope units over the mapping period, together with the landslide frequency.}
	\label{fig:fig1}
	\centering
\end{figure}

Specifically, their polygonal inventory reports yearly landslide occurrences for each year from 1988 to 2018 in the part of Nepal where most landslides occur (see Figure~\ref{fig:fig1}). Interestingly, their mapping efforts excluded any seismic or anthropogenic influences, being a unique type of inventory within the Nepali landscape and at the global level.   
This landslide inventory is the first of its kind and, despite ignoring the exact timing of landslides within each year, it has the potential to provide extensive information for modelling the year-by-year evolution of landslides in a spatiotemporal fashion. Landslides were mapped based on satellite imagery. Therefore, with such a large time span, landslides had to be recognised in different scenes with varying resolutions. This may induce some bias in the quality \cite{guzzetti2012landslide} and completeness \cite{TanyasLombardo2020} of the data. However, \citeA{jones202130} kept the same mapping procedure across time and supporting data, thus reducing any potential bias to the minimum \cite{jones202130}. Furthermore, the yearly characteristics of the inventory highlight the landscape evolutionary process in terms of newly activated and reactivated landslides.

\section{Mapping Units}
\label{sec:Mapping Units}

More than thirty years ago, \citeA{carrara1983multivariate,carrara1988drainage} proposed a terrain unit capable of closely reflecting landscape morphology in data-driven statistical models. These were named slope units (SU, hereafter), representing the physiographic space bounded between ridges and streamlines. Therefore, in this work, we opted to use a SU partition, which is better suited than one based on grid cells when modelling landslides over large spatiotemporal domains. To construct the SU partition automatically, we used the function \textit{r.slopeunits} \cite{alvioli2016automatic}, the most common software used for SU delineation. The \textit{r.slopeunits} function implements a customized version of the watershed algorithm and requires a Digital Elevation Model (DEM), as well as a suite of input parameters that are used to converge to a deterministic solution. 
These input parameters are: \textit{i}) Initial flow accumulation threshold (\emph{thresh}); \textit{ii}) Circular variance (\emph{cvmin}); \textit{iii}) minimum SU surface area (\emph{areamin}); \textit{iv}) maximum SU surface area (\emph{maxarea}); \textit{v}) reduction factor (\emph{rf}); \textit{vi}) the threshold value for the cleaning procedures (\emph{cleansize}). Their respective meanings are detailed in \citeA{alvioli2016automatic}. Because the choice of input parameters will determine the resulting SU partition, in this work, we tested several input parameter combinations, whose value ranges and our final setting are reported in Table~\ref{Table:suval}. 

\begin{table}[t!]
	\caption{Table showing the input parameters to generate the SU partition.}
 \label{Table:suval}
	\begin{tabular}{@{}lllll@{}}
		\hline
		Parameter                                                                     & Selected Value & Minimum Range & Maximum Range \\ \hline
		Minimum Area                                                                  & 40\,000       & 20\,000                & 50\,000                \\
		Circular Variance                                                             & 0.4         & 0.2                  & 0.6                  \\
		Clean Size                                                                    & 20\,000       & 10\,000               & 30\,000                \\
		Reduction Factor                                                              & 10          & 10                   & 40                   \\
		\begin{tabular}[c]{@{}l@{}}Initial Flow\\ Accumulation Threshold\end{tabular} & 80\,000       & 60\,000                & 100\,000                \\
		Maximum Iteration                                                             & 15          & 15                   & 15                  
	\end{tabular}
\end{table}
Aside from input parameters, the DEM we used comes from the 30~m SRTM \cite{berry2007near} produced by the United States Geological Survey (USGS), which has already been hydrologically corrected. Note that the latest version of \textit{r.slopeunits} that we used is capable of filtering out flat regions, as they are irrelevant from a landslide perspective. 

\begin{figure}[t!]
        \centering
	\includegraphics[scale=0.9]{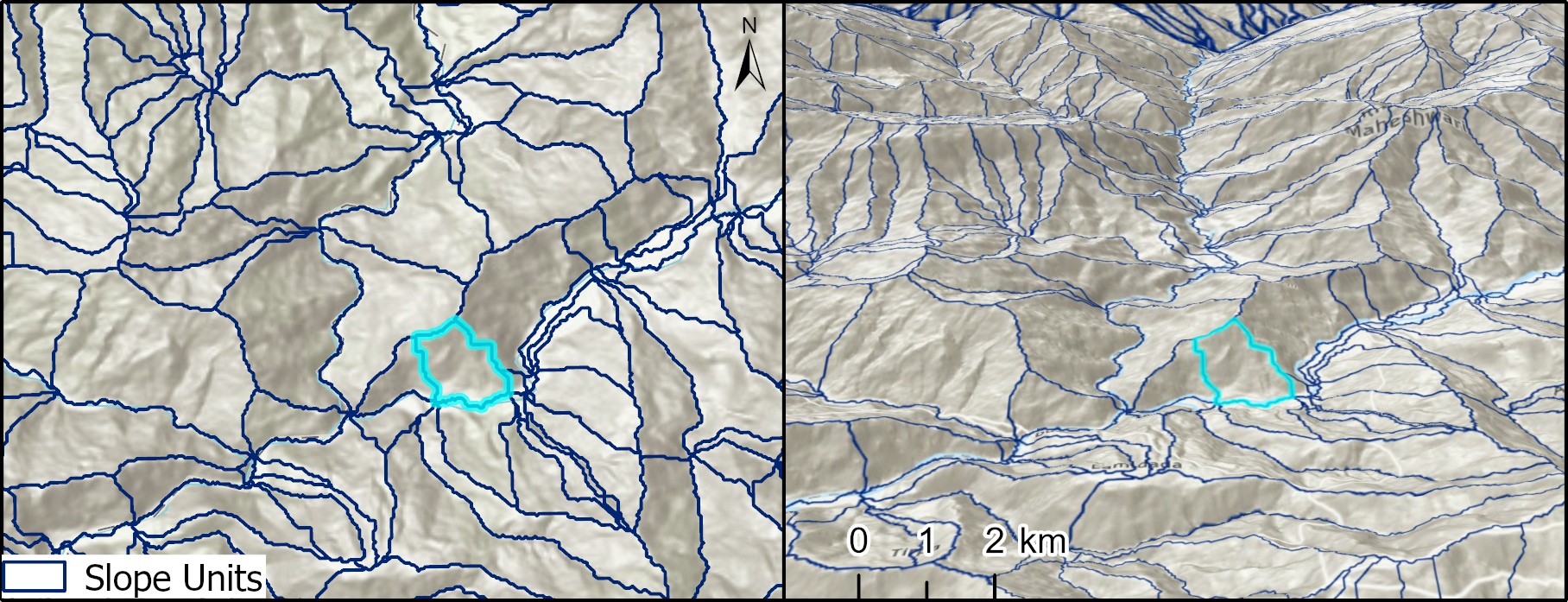}
	\caption{Samples of SUs in 2D map with shaded relief (left) and 3D shaded relief (right) with camera angle of  $30 ^\circ$ from south to north. Highlighted SUs in both maps show the same location.}
 \label{fig:fig2}
	\centering
\end{figure}

Figure~\ref{fig:fig2} shows the details of our SU partition. Specifically, the left and right panels present a 2D and 3D rendering, respectively, of a subdomain with SU boundaries highlighted in blue. For each SU and each year, we have then aggregated the landslide inventory with landslide polygons and estimated the total area density of landslides, defined as the ratio of the total area of slope failure over the entire SU area, over a spatiotemporal range of our study. The area density is used thereafter as a proxy for the landslide size, used to estimate the landslide intensity.

\begin{figure}[t!]
        \centering
	\includegraphics[scale=0.6]{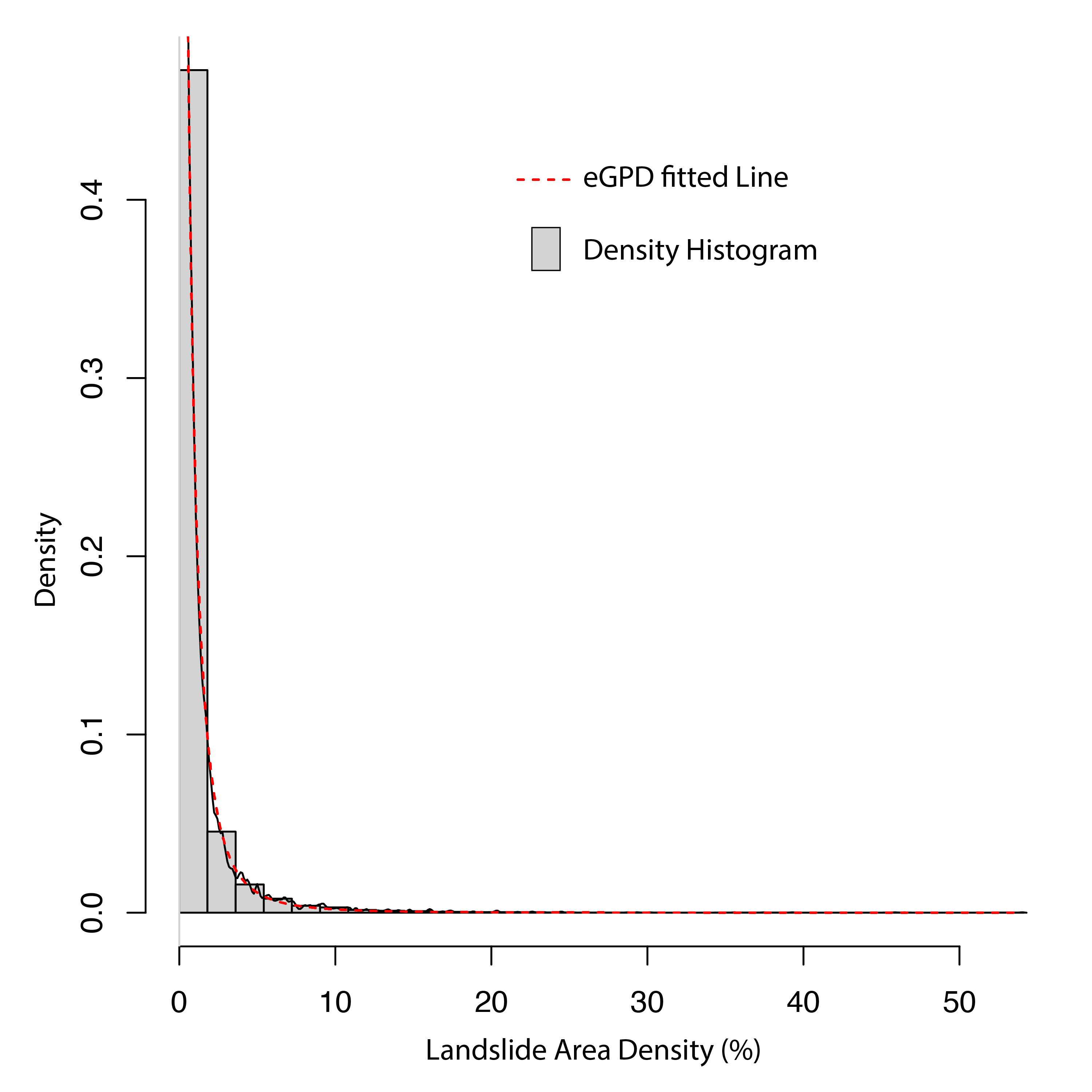}
	\label{fig:distribution}
	\caption{Empirical distribution of the landslide area density (histogram) and the best-fitting extended Generalised Pareto Distribution (eGPD) probability density function (dotted red).}
	\centering
\end{figure}

To better understand the distribution of the landslide area density, which is crucial to designing a sensible landslide hazard model, we then compared the (unconditional) empirical distribution of the landslide area density (by pooling data over both space and time) with a suite of probability models. The best-fitting model ended up being the extended Generalised Pareto Distribution (eGPD) \cite{papastathopoulos2013extended,naveau2016modeling}; its fit is displayed in  Figure~\ref{fig:distribution}. We can observe that the eGPD indeed fits the data very well, all the way from low to high quantiles. The eGPD is an extension of the Generalized Pareto Distribution (GPD), which is a continuous probability distribution frequently employed to characterise the extreme tails of another distribution \cite <see,>[]{pickands1975statistical,castillo1997fitting,naveau2016modeling}. While the classical GPD is defined in terms of scale ($\sigma$) and shape ($\xi$) parameters, the eGPD includes one more parameter ($\kappa$) to model the lower-tail region of the distribution, thus enabling it to describe extreme and bulk behaviours jointly via a single model \cite{cisneros2023deep}. As a parsimonious yet relatively flexible parametric distribution respecting extreme-value theory in its tails, this distribution is suitable for modeling the landslide area density. The methods section (Section~\ref{sec:methods}) will further describe why and how we can use the eGPD to model and predict the landslide hazard. 

\section{Predictor Set: Static and Dynamic Properties}
\label{sec:predictors}

To explain the distribution and size of landslides across space and time, we opted for a large predictor set. All properties capable of exerting a given influence on the genesis and evolution of landslides vary with time. However, in a relatively short time window, some environmental parameters can be approximated to be temporally stationary, whereas others exhibit a much higher rate of temporal variation. For this reason, in the remainder of this manuscript, we will refer to these two predictor types as static and dynamic predictors, respectively. 

Table \ref{Table:Input} lists both predictor types, further divided into five categories according to their nature and the information they convey to the model. Those categories mainly represent morphometrics, vegetation, triggering factors, soil properties, and geology. The mean value and standard deviation of these variables within each SU are used in the modelling process. Furthermore, we recall that since landslides are here observed yearly, dynamic predictors must also be aggregated on a yearly scale (even if observed at a higher temporal resolution).

\begin{table}[t!]
	\caption{Predictors used in the model, with dynamic predictor categories highlighted in \textbf{bold}.}
 \label{Table:Input}
	\begin{tabular}{@{}lcll@{}}
		\hline
		\textbf{Parameter}                                                                          & \multicolumn{1}{l}{\textbf{Category}}                                                             & \textbf{Data Source}                                                            & \textbf{Reference} \\ \hline
		Elevation Mean (meters)                                                            & \multicolumn{1}{l}{\multirow{10}{*}{\begin{tabular}[c]{@{}l@{}}Morpho- \\ Metrics\end{tabular}}}                                                                 & \multirow{10}{*}{\begin{tabular}[c]{@{}l@{}}SRTM \\ DEM (30 m)\end{tabular}}    &\multirow{10}{*}{\scriptsize\citeA{farr2000shuttle}}\\
		Elevation Standard Deviation                                                       &                                                                                                   &                                                                                 &                    \\
		Slope Mean (degrees)                                                               &                                                                                                   &                                                                                 &                    \\
		Slope Standard Deviation                                                           &                                                                                                   &                                                                                 &                    \\
		Aspect Mean (radians)                                                              &                                                                                                   &                                                                                 &                    \\
		Aspect Standard Deviation                                                          &                                                                                                   &                                                                                 &                    \\
		Horizontal Curvature Mean                                                          &                                                                                                   &                                                                                 &                    \\
		\begin{tabular}[c]{@{}l@{}}Horizontal Curvature  Standard Deviation\end{tabular} &                                                                                                   &                                                                                 &                    \\
		Vertical Curvature Mean                                                            &                                                                                                   &                                                                                 &                    \\
		\begin{tabular}[c]{@{}l@{}}Vertical Curvature  Standard Deviation\end{tabular}   &                                                                                                   &                                                                                 &                    \\ \hline
		NDVI Mean                                                                          & \multicolumn{1}{l}{\multirow{2}{*}{\textbf{Vegetation}}}                                                   & \multirow{2}{*}{Landsat 5,7}                                                    &    \multirow{2}{*}{\scriptsize\citeA{tucker1979red}}                \\
		NDVI Standard Deviation                                                            & \multicolumn{1}{l}{}                                                                              &                                                                                 &                    \\ \hline
		Precipitation Max (mm)                                                             & \multicolumn{1}{l}{\multirow{3}{*}{\begin{tabular}[c]{@{}l@{}}\textbf{Triggering} \\ \textbf{Factors}\end{tabular}}} & \multirow{3}{*}{CHIRPS}                                                         &     \multirow{3}{*}{\scriptsize\citeA{funk2015climate}}               \\
		Precipitation Mean (mm)                                                            & \multicolumn{1}{l}{}                                                                              &                                                                                 &                    \\
		\begin{tabular}[c]{@{}l@{}}Precipitation  Standard Deviation\end{tabular}        & \multicolumn{1}{l}{}                                                                              &                                                                                 &                    \\ \hline
		Clay Content (gm/cu.cm)                                                            & \multicolumn{1}{l}{\multirow{2}{*}{\begin{tabular}[c]{@{}l@{}}Soil \\ Properties\end{tabular}}}   & \multirow{2}{*}{\begin{tabular}[c]{@{}l@{}}Soilgrids\end{tabular}} &  \multirow{2}{*}{\scriptsize\citeA{hengl2017soilgrids250m}}                  \\
		Organic Content (gm/cu.cm)                                                         & \multicolumn{1}{l}{}                                                                              &                                                                                 &  
		\\ \hline
		Geological categories                                                           & \multicolumn{1}{l}{\multirow{1}{*}{\begin{tabular}[c]{@{}l@{}}Geology \end{tabular}}}   & \multirow{1}{*}{\begin{tabular}[c]{@{}l@{}}USGS\end{tabular}} &    \multirow{1}{*}{\scriptsize\citeA{wandrey1998maps}}                                
	\end{tabular}
\end{table}

To compute all the variables listed in 
Table~\ref{Table:Input}, we used multiple approaches based on the spatial and temporal resolution of the input datasets.
Regarding static (morphometric, soil-related, geological) predictors, we simply computed their mean and standard deviation per SU. Dynamic predictors (of a time-varying nature), however, were available at different spatial and temporal resolutions, so a special treatment was required for each of them. The high spatial resolution of NDVI allowed us to compute the mean value and standard deviation per SU from available NDVI scenes in a year. The coarse resolution of CHIRPS data constrained us to assign to any SU falling within a $5 \times 5$~km grid cell the actual precipitation value, aggregated as: \textit{i}) the daily maximum in a year; \textit{ii}) the daily mean in a year; and \textit{iii}) the daily standard deviation in a year. All these operations were directly implemented in Google Earth Engine to speed up the whole preprocessing pipeline, and the code is shared in GitHub (\url{https://github.com/ashokdahal/extremevaluelandslides}).

\begin{figure}[t!]
	\includegraphics[scale=0.8]{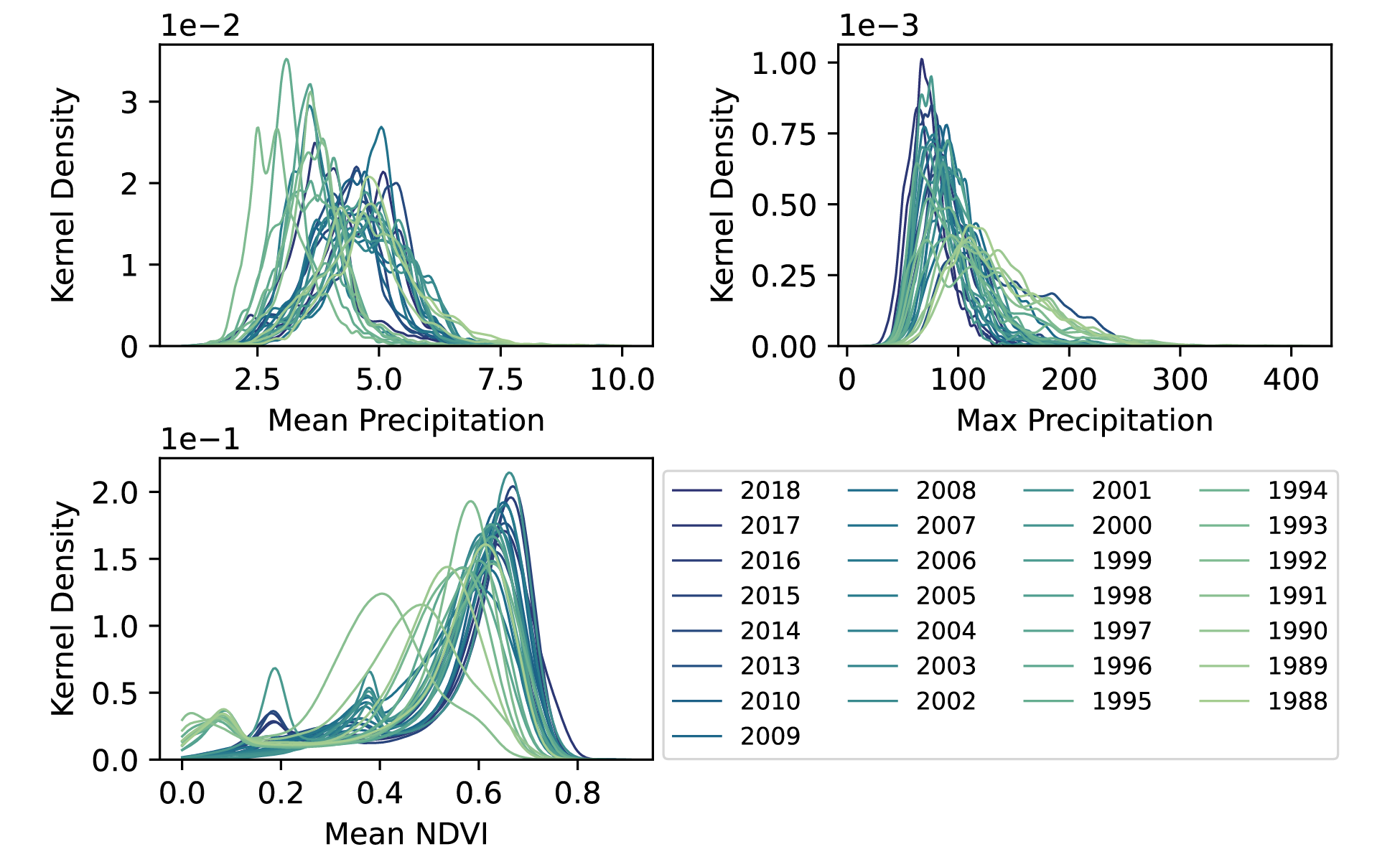}
	\label{fig:fig3}
	\caption{Distribution of Mean Precipitation (mm/day; top left), Maximum Precipitation (mm/day; top right) and Mean NDVI (unitless; bottom left) over 30 years pooling the data from all SUs together, obtained using Kernel Density Estimation.}
	\centering
\end{figure}

To provide an overview of the constructed dynamic predictors, which change yearly, the mean and maximum precipitation and mean NDVI empirical distribution are displayed in Figure~\ref{fig:fig3}. We can observe that the spatial distribution of all dynamic variables has a similar yearly trend. The distribution of the mean precipitation in recent years (2009--2018) clusters together, peaking within the range of 3--6~mm/day. Similarly, the maximum precipitation for all years has a similar spatial distribution, with a right heavy tail. Moreover, NDVI shows peaks in the range of 0.4--0.6 for data before 1999; this is likely due to spectral and spatial resolution limitations in old sensors rather than significant environmental changes. This was later solved (see peaks clustered at around 0.65 from early 2000 onwards) by the improvement in both spatial and spectral resolutions of Landsat-7 (launched in 1999). 

\section{Methods}
\label{sec:methods}
\subsection{Hazard Definition and Modelling}
The landslide hazard was initially qualitatively defined by Varnes and the International Association for Engineering Geology commission on landslides and other mass movements in 1984 as \textit{``the probability of occurrence within a specified period and a given area of a potentially damaging phenomenon"}. 
This definition was later refined by \citeA{guzzetti1999landslidehazard} to include the intensity of the event. 
As a landslide can either be present or absent at a given point in space and time, we can represent it by the binary random variable $L(s,t)$, which takes value one if a landslide (or more than one) triggers at spatial location (here, a SU) $s$ and time (here, a year) $t$, and is equal to zero otherwise. Similarly, the intensity of a landslide should reflect its ``destructiveness'' (e.g., a function of its ``size'' in terms of the displaced mass, its volume or planimetric extent); in our study, we express landslide size in terms of the area density (i.e., the fraction of each SU affected by a landslide), represented by a generic nonnegative random variable $A(s,t)$ (for ``area''), also indexed by space and time.
Obviously, $L(s,t)$ and $A(s,t)$ are not independent of each other, given that $L(s,t)=0$ if and only if $A(s,t)=0$. Mathematically, the overall landslide hazard can thus be quantatively expressed from the above definition as the joint probability
$${\rm Pr}(\cup_{(s,t)\in\mathcal{X}}\{L(s,t)=1,A(s,t)\geq a_q\}),$$
where $\mathcal{X}$ is a specified spatio-temporal domain (spatial area and time period) of interest, and $a_q>0$ is some critical landslide size of ``severity'' $q\in(0,1)$; there are different ways to define $a_q$, for example as the $(100\times q)$-percentage of the area of SU failure, or as the $q$-quantile from the \emph{unconditional distribution} of observed landslide sizes (i.e., pooled over space and time). Both have pros and cons, but in this work, we choose the latter because $q$ can be directly interpreted as an overall landslide size level. From a practical mitigation perspective, rather than computing the \emph{overall} landslide hazard for an entire domain and time period, it is often more helpful to estimate the \emph{local} hazard at each specific location $s$ and time $t$, and to understand how it relates to the main triggering driver (e.g., heavy precipitation) summarised in a vector $\boldsymbol{X}_D(s,t)$ of features, and other important covariates summarised in a vector $\boldsymbol{X}_O(s,t)$. Therefore, inspired by the above definition, we shall here formalise the $q$-level (local) landslide hazard at location $s$ and time $t$ as 
\begin{equation}
\label{eq:hazard2}
h_q(s,t)=p(s,t)\times i_q(s,t),
\end{equation}
where $p(s,t)={\rm Pr}\{L(s,t)=1\mid \boldsymbol{X}_D(s,t)=\boldsymbol{x}_D(s,t),\boldsymbol{X}_O(s,t)=\boldsymbol{x}_O(s,t)\}$ is the landslide susceptibility (i.e., the landslide occurrence probability given the main driver and covariates) and 
$i_q(s,t)={\rm Pr}\{A(s,t) \geq a_q\mid {L(s,t)=1}, \boldsymbol{X}_D(s,t)=\boldsymbol{x}_D(s,t),\boldsymbol{X}_O(s,t)=\boldsymbol{x}_O(s,t)\}$ is the landslide intensity expressed as the probability that the landslide size exceeds the chosen threshold $a_q$ given that a landslide has occurred, and given the main driver and covariates. While this formalism superficially resembles the probabilistic hazard definition introduced by \citeA{guzzetti2005probabilistic}, there are key differences with their approach that we highlight further below. The choice of $q\in(0,1)$ should, in principle, depend on the nature/vulnerability/cost of the elements at risk in each SU, as this may determine the local hazard level that can be tolerated or accepted. 
In practice, $q$ should thus ideally be set according to the Government rules and guidelines for the different infrastructures and be SU-specific in a potentially time-varying manner. Here, however, for simplicity in the absence of such governmental information, we keep $q$ (and thus, $a_q$) fixed over space and time, but we showcase the landslide hazard for different critical levels, namely with $q=5\%,50\%,95\%$.   

In order to account for the frequency of the landslide trigger event in the hazard computation and to assess how the landslide hazard changes according to different possible future scenarios, we additionally build a statistical extreme-value model for the main triggering driver (here, precipitation), from which we compute long-term return levels for each spatial location (i.e., each SU). Specifically, assuming temporal stationarity, we estimate the vector $\boldsymbol{\rm RL}_P(s)$ of return levels for each of the relevant components of the feature vector $\boldsymbol{X}_D(s,t)=\boldsymbol{x}_D(s,t)$, defined componentwise as the level that is expected to be exceeded once by the selected variable in the period $P$ (e.g., with $P=5,10,15,20$ years) at location $s$. 
Then, for a given return period $P$, we plug the estimated return levels (along with the other static covariates and the time-averaged NDVI covariate) into the fitted local hazard model $h_q(s,t)$ in \eqref{eq:hazard2} by replacing the observed triggering driver variables $\boldsymbol{X}_D(s,t)$ with $\boldsymbol{\rm RL}_P(s)$.
This allows us to estimate the ``hypothesised'' $q$-level spatial landslide hazard, $h_{q;P}(s)$ that would be obtained should each summary feature of the triggering driver reach its $P$-year return level everywhere in space. 
Such hypothesised scenario-based hazard maps go beyond the approach advocated by \citeA{guzzetti2005probabilistic} and provide useful information on the potential impact of future landslides for periods of time of specified lengths.

We recall that in our study, landslide data are observed yearly and are spatially aggregated at the SU level, so $L(s,t)$ indicates the presence/absence of at least one landslide per SU $s$ during year $t$, and similarly, $\boldsymbol{X}_D(s,t)$ and $\boldsymbol{X}_O(s,t)$ denote features of the main triggering driver and other covariates summarised yearly at the SU resolution. However, the same concepts and methodology could be formulated with data at different resolutions (e.g., with data observed daily or monthly at a pixel resolution) with only slight adjustments. 
We now give more details on the modelling of each component (landslide susceptibility, landslide intensity, triggering driver frequency through return level estimation).




\subsubsection{Susceptibility Component, $p(s,t)$}

The binary variable $L(s,t)$, representing the presence/absence of landslides per SU $s$ at time $t$, can be modelled using a Bernoulli distribution, i.e.,
\begin{equation}
\label{eq:susceptibility}
L(s,t)\sim {\rm Ber}(p(s,t)),
\end{equation}
where the probability $p(s,t)\in(0,1)$ is the landslide susceptibility that we seek to estimate. 
To this end, we express the susceptibility $p(s,t)$ in terms of the main triggering driver variables $\boldsymbol{X}_D(s,t)=\boldsymbol{x}_D(s,t)$ and the other covariates $\boldsymbol{X}_O(s,t)=\boldsymbol{x}_O(s,t)$ through a flexible deep-learning model, i.e., a potentially highly nonlinear function estimated from the data (see Section~\ref{sec:experiment}). 
We note that while the driver and covariates may change over space and/or time, the functional relationship between them and the landslide susceptibility, $p(s,t)$, is assumed to be constant.
This assumption allows us to use our model to predict the susceptibility under new triggering driver and covariate values (e.g., under more extreme trigger conditions), but it also requires a leap of faith as the ability of a model to extrapolate to yet-unseen conditions cannot be easily verified in practice.

\subsubsection{Intensity Component, $i_q(s,t)$}
The landslide intensity is defined as the probability that the landslide size (here, expressed in terms of landslide area density), $A(s,t)$, exceeds a certain critical threshold $a_q$ conditional on a landslide occurrence (i.e., conditional on $A(s,t)>0$) in a given SU $s$ and specific year $t$. The nonnegative random variable $A(s,t)$ has a right-skewed distribution and often exhibits a relatively heavy upper tail (recall Figure~\ref{fig:distribution}) despite having a finite upper bound due to the physical constraint that the area of each SU is always finite. 
We therefore seek a flexible probability distribution that can suitably capture these features. 
Moreover, because we shall compute the landslide hazard at low-to-high critical levels, e.g., $q=5\%,50\%,95\%$, we need a distribution that can accurately capture both the bulk behaviour and the lower/upper tails of the landslide area density. To this end, joint bulk-tail models motivated by extreme-value theory are thus ideally placed.

One particularly interesting model is the so-called extended generalised Pareto distribution (eGPD; \citeA{papastathopoulos2013extended,naveau2016modeling}), which is controlled by three parameters. The corresponding distribution function can be written as 
\begin{equation}
\label{eq:eGPD}
F_{\rm eGPD}(x;\kappa,\sigma,\xi)=\left\{1-(1+\xi x/\sigma)^{-1/\xi}\right\}^\kappa,\quad x>0,
\end{equation}
where $\kappa>0$ is the lower-tail shape parameter, $\sigma>0$ is the scale parameter, and $\xi>0$ is the upper-tail shape parameter. With $\xi>0$, the eGPD is heavy-tailed and the tail heaviness increases as $\xi$ increases.  
When $\kappa=1$, the eGPD reduces to the widely-used generalised Pareto distribution (GPD), which is the only possible limiting tail behaviour according to extreme-value theory (\citeA{davison2015statistics}). 
The eGPD has not only lower and upper tail behaviors that comply with extreme-value theory, but it also has a smooth transition in between (i.e., in the bulk).
This allows one to use the eGPD for modeling the full range of the data (from low to high quantiles), while bypassing the specification of low/high thresholds as commonly required in extreme-value analyses. 
This parsimonious yet flexible distribution has thus been used extensively for the statistical modelling of natural hazards, including precipitation (\citeA{naveau2016modeling}), wildfires (\citeA{cisneros2023deep}), and even landslides (\citeA{yadav2022joint}). 
However, it is the first time that it is used in a full landslide hazard assessment context, particularly with the proposed deep regression modelling approach.
While the eGPD does not have a finite upper endpoint (required given that the maximum possible area density is equal to $100\%$), it would be possible to use a truncated eGPD instead, whereby the model \eqref{eq:eGPD} would be replaced by $F_{\rm eGPD}(x;\kappa,\sigma,\xi)/F_{\rm eGPD}(1;\kappa,\sigma,\xi)$ for $x\in[0,1]$; however, as seen in Figure~\ref{fig:fig3}, the distribution of area densities has almost all its mass in the interval $[0,20]\%$; therefore, $F_{\rm eGPD}(1;\kappa,\sigma,\xi)\approx1$ in our case, which means that the truncated and untruncated models are almost equivalent to each other, and we have found that the untruncated eGPD in \eqref{eq:eGPD} still provides an excellent fit in practice, especially if we do not used the fitted model to extrapolate too far into the upper tail.

Assuming that the scale $\sigma$ depends on the main triggering driver variables $\boldsymbol{X}_D(s,t)$ and the other covariates $\boldsymbol{X}_O(s,t)$, but keeping the shape parameters, $\kappa$ and $\xi$, constant over space and time to get a parsimonious representation, we thus model positive landslide sizes using \eqref{eq:eGPD} as
$$A(s,t)\mid A(s,t)>0\sim F_{\rm eGPD}(x;\kappa,\sigma(s,t),\xi),$$
from which we easily derive the landslide intensity as 
\begin{align}
i_q(s,t) &= 1-F_{\rm eGPD}(a_q; \kappa,\sigma(s,t),\xi)\nonumber\\
&=1-\left\{1-\bigg(1+\frac{\xi a_q}{\sigma(s,t)}\bigg)^{-1/\xi}\right\}^\kappa.\label{eq:intensity}
\end{align}

Similarly to the susceptibility component, we model the scale $\sigma(s,t)$ in terms of a potentially highly nonlinear function of $\boldsymbol{X}_D(s,t)$ and $\boldsymbol{X}_O(s,t)$ through a flexible deep-learning model (see Section~\ref{sec:experiment}). 
As before, we assume that the functional relationship linking $\boldsymbol{X}_D(s,t)$ and $\boldsymbol{X}_O(s,t)$ with the scale $\sigma(s,t)$ does not change with location $s$, time $t$, or trigger/covariate values. This allows us to extrapolate the intensity to new trigger/covariate conditions as well as to the unobserved (potentially extreme) severity levels $q$.

\subsubsection{Frequency Component: Return levels for the triggering driver}
\label{eq:freq}
The major problem when assessing landslide hazard is that it is a ``cascading event'' \cite{tilloy2019review} and therefore, one should take the frequency of the landslide trigger into account --- an important aspect that has rarely been considered previously in the data-driven landslide modelling literature. 
In other words, the occurrence of a landslide mainly depends on whether sufficient precipitation has occurred (for rainfall-induced landslides) or whether sufficient ground shaking has occurred (for earthquake-induced landslides).
Hence, in practice, landslide hazard is only meaningful as a concept if it involves the frequency of the trigger (extreme rainfall/shaking) in its calculation. 
While landslides may sometimes occur as ``conditional hazards'' \cite{tilloy2019review}, such as landslides occurring due to long-term preconditioning factors rather than being due to an external trigger, these landslide types are out of the scope of this paper. 
So, rather than asking when or how frequently landslides of a given intensity occur (as, e.g., done by \citeA{guzzetti2005probabilistic}), it is more natural to ask how frequently a specific triggering event (rainfall/shaking) reaches extreme levels, assuming everything else remains constant (which is justified when the study area does not go under rapid change), and how such extreme conditions may, in turn, modify the landslide hazard. 

In this work, we have precisely defined the hypothesised landslide hazard to reflect this notion: namely, what the hazard would be should the trigger reach a specified extreme level. 
To this end, we compute long-term return levels for summary features of the main triggering driver (in the vector $\boldsymbol{X}_D(s,t)=\boldsymbol{x}_D(s,t)$) by building separate extreme-value models for each of the relevant components of $\boldsymbol{x}_D(s,t)$.

Specifically, we build two separate deep-learning-based eGPD models that resemble the landslide intensity model, but fitted to both annual maximum and annual mean precipitation data; we treat the annual precipitation standard deviation differently (as described at the end of this section). 
For simplicity, with some slight abuse of notation, let $X_D(s,t)=x_D(s,t)$ denote either the annual maximum precipitation or the annual mean precipitation.  
Note that because we here model precipitation data at the yearly scale, we avoid complications with temporal dependence and non-stationarity.
Moreover, as zero annual precipitation is never observed, we can neglect the point mass at zero. 
Assuming that the data are independent and identically distributed over the years, our model may be expressed as 
$$X_D(s,t) \sim F_{\rm eGPD}(x; \tilde\kappa,\tilde\sigma(s),\tilde\xi),$$
where $\tilde\kappa>0,\tilde\xi>0$ are constant lower-tail and upper-tail shape parameters, respectively, and $\tilde\sigma(s)>0$ is now assumed to only vary across space through a similar deep-learning formulation, yet without the use of covariates.

We note that while the generalised extreme-value (GEV) distribution is a more common choice to model annual maxima, the eGPD is a valid model for block maxima from eGPD data. 
In other words, if daily rainfall follows an eGPD, then annual rainfall maxima will also follow an eGPD, yet with a different lower-tail $\kappa$ parameter. 
Moreover, it can be shown that the eGPD is close to a GEV distribution when $\kappa\to\infty$, and unlike the GEV distribution, it is always supported over $(0,\infty)$, which makes sense for precipitation observed on any scale. 
Our model is, therefore, sensible for both annual maximum and mean precipitation. 
Once the models are trained, estimated parameters can be plugged into the eGPD quantile function to compute return levels of the maximum and mean precipitation. 
Specifically, the $P$-year return level of $X_D(s,t)$ at location $s$ (i.e., the level that is expected to be exceeded once in $P$ years at location $s$) is obtained as 
$${\rm RL}_P(s)=F_{\rm eGPD}^{-1}(1-1/(Pn_y);\tilde\kappa,\tilde\sigma(s);\tilde\xi)={\tilde\sigma(s)\over\tilde\xi}\left[(1-\{1-1/(Pn_y)\}^{1/\tilde\kappa})^{-\tilde\xi}-1\right],$$
where $n_y$ is the number of independent observations of $X_D(s,t)$ within a year. 
Since our precipitation data are modelled on an annual basis, we here set $n_y=1$.

Regarding the annual precipitation standard deviation (which is also an element of the feature vector $\boldsymbol{X}_D(s,t)=\boldsymbol{x}_D(s,t)$), it does not make sense to compute return levels for it. Therefore, we used an ad-hoc approach, whereby the most likely precipitation analogue was identified from the observed yearly data, from which the standard deviation was then extracted. More precisely, for each considered return period, we identified a year with the closest mean and maximum precipitation to that of the estimated annual mean and maximum precipitation return levels. The year with the most similar precipitation analogue was identified by minimizing the Euclidean distance between observed and modelled mean and maximum precipitation return levels across all years.

Return level estimates for the annual maximum and mean precipitation, as well as the annual precipitation standard deviation of the closest analogue, were then combined into the vector $\boldsymbol{\rm RL}_P(s)$, plugged into \eqref{eq:hazard2} to compute the hypothesized landslide hazard $h_{q;P}(s)$.

\subsection{Hazard Computation and Future Projections Based on Climate Change Scenarios}

One of the main reasons to build a landslide hazard model is to understand the threat posed by landslides, which is often dependent on the intensity of triggering events.
Landslide hazard posed by different intensities of triggering events (represented by return levels for various return periods) can be used to understand the landslide risk and mitigate it. 
Thus, to develop a test case for our landslide hazard model, we designed multiple hypothesised (spatially-varying) rainfall scenarios corresponding to the $5$, $10$, $15$, and $20$-year return levels, estimated based on the eGPD model presented in Section~\ref{eq:freq} using the CHRIPS rainfall data recorded from 1988--2018.
Such designed rainfall scenarios allow us to understand how landslide hazard changes depending on the intensity of the rainfall trigger, assuming everything else remains constant.
Using precipitation data up to 2018 allowed us to assess the landslide hazard under different scenarios, by varying the rainfall return period. 

While our model gives us access to the full landslide hazard distribution for different exceedance probability thresholds ($q$), we have computed the hazard for three different failure scenarios corresponding to the $5\%$, $50\%$, and $95\%$ of the observed landslide area distribution in the past 30 years. 
The associated landslide area density per SU are $0.017\%$, $0.423\%$ and $4.245\%$, respectively.

Notably, the precipitation patterns can vary over time due to climate change. 
This may result in extended periods of extreme drought and intense precipitation. 
In turn, this should affect the landslide hazard. 
To understand how landslide hazard may change in the future under different climate change scenarios, we repeated our analysis based on two climate change pathways, namely, the middle pathway (SSP245) and the fossil-fueled development pathway (SSP585). 
The projected precipitation data were obtained from ensembled climate change projections of the Coupled Model Intercomparison Project (CMIP6) \cite{taylor2012overview}, downscaled and bias corrected by \citeA{thrasher2022nasa}. 
In this research, we only used the models downscaled by \citeA{thrasher2022nasa} from the original CMIP6 database because including other models would require significant effort on bias correction and downscaling, which is beyond our capabilities. 
The time period of 2020--2100 was considered to design the hypothesised rainfall scenarios (i.e., to compute rainfall return levels). 
Therefore, the predicted hazard results can potentially reflect the average landslide hazard up to the end of the century. 
Notably, climate projection data are usually produced at a coarse resolution of $30\times30$ km, making their spatial distribution biased and less relevant for a SU-based model. 
Therefore, the landslide hazard projections displayed in this research should be treated as an indication of the model's usability and interpreted with care. 

\subsection{Deep-Learning Framework}
\label{sec:experiment}
\subsubsection{Neural Network Architecture}
The proposed landslide hazard model has two probability distributions to learn: namely, Bernoulli and eGPD, each having a parameter (i.e., $p(s,t)$ and $\sigma(s,t)$, respectively) varying nonlinearly over space and time through a complex interaction of multiple static and dynamic covariates. 
In addition, the eGPD model has two extra parameters held constant over space and time (namely $\kappa$ and $\xi$). 
The joint estimation of all parameters given the input covariate data $\boldsymbol{X}_D(s,t)$ and $\boldsymbol{X}_O(s,t)$ is performed efficiently through a shared deep-learning architecture with two output layers. The architecture is ``shared'' here since we use a single deep-learning model (with the same underlying parameters) to generate the two outputs. 
This shared architecture is a natural framework to use here as the parameters $p(s,t)$ and $\sigma(s,t)$ are expected to be correlated, and it also makes the neural network more parsimonious while retaining high flexibility. 
More specifically, the designed model consists of 16 artificial neural network blocks, each consisting of a dense layer with a width of 64 neurons followed by a dropout layer, a batch normalisation layer, and a rectified linear unit (ReLU) activation function (for further details on those layers, see \citeA{dahal2023explainable}).
The dropout and batch normalisation layers prevent the model from overfitting \cite{srivastava2014dropout}.
At the same time, the ReLU activation function allows the model to be very flexible and to capture complex nonlinear patterns in the data. 
The two output layers are then passed through different final activation functions to constrain their range within the acceptable domain of the predicted parameters. Specifically, the $p(s,t)$ parameter is passed through a sigmoid function to constrain it within the interval $(0,1)$, while the $\sigma(s,t)$ parameter is passed through a ReLU layer to enforce nonnegativity, i.e, $\sigma(s,t)\geq0$.

\subsubsection{Loss Function}
To train the model, we designed the loss function as a weighted summation of the weighted binary cross-entropy loss, which is a popular loss function for imbalanced binary data in machine learning (similar to a weighted negative Bernoulli log-likelihood), and the negative log-likelihood of the eGPD distribution \cite{papastathopoulos2013extended,cisneros2023deep}.
Precisely, assume that we have observed landslide presence/absence and landslide size (areal density) data at $n$ space-time locations, denoted by $(s_i,t_i)$, $i=1,\ldots,n$, and let $\ell_i$ and $a_i$ denote the realised values of $L(s,t)$ and $A(s,t)$ at $(s_i,t_i)$. Similarly, let $p_i(\boldsymbol{\varphi})$ and $\sigma_i(\boldsymbol{\varphi})$ denote the values of $p(s_i,t_i)$ and $\sigma(s_i,t_i)$, respectively, for each $i=1,\ldots,n$, which depend on the unknown vector $\boldsymbol{\varphi}$ containing all the learnable neural network parameters (i.e., so-called ``weights'' and ``biases''). 
In practice, rather than using the full training data at each optimisation iteration, we often use a smaller (randomly selected) batch $B_N\subset\{1,\ldots,n\}$ of size $N<n$, which makes it more computationally affordable.
Then, the combined loss function that we minimised can be expressed as
\begin{equation}
\label{eq:loss}
\begin{split}
\Psi(\boldsymbol{\varphi},\kappa,\xi)= \gamma \bigg[-\sum_{i\in B_N} \left\{0.9\ell_i \log(p_i(\boldsymbol{\varphi}))+0.1(1-\ell_i)\log(1-p_i(\boldsymbol{\varphi}))\right\}
\bigg]+\\
(1-\gamma) \bigg[ - \sum_{i\in B_N} \ell_i\log\big( f_{\rm eGPD}(a_i; \kappa,\sigma_i(\boldsymbol{\varphi}),\xi)\big) \bigg]
\end{split}
\end{equation}
where $\gamma\in(0,1)$ is a fixed regularisation hyperparameter controlling the balance between the Bernoulli and eGPD parts, and $f_{\rm eGPD}$ is the eGPD density function obtained as the first derivative of the distribution function in \eqref{eq:eGPD}. In this case, $\gamma$ was tuned in the range 0.3--0.7 with a step size of 0.05, and we identified 0.5 as the best value, but this might change with each dataset. 
Thus, tuning the $\gamma$ parameter is always recommended. 
Furthermore, in the first part of the equation, the weighted binary cross-entropy loss has a weight of $0.9$ and $0.1$ for 
landslide presences and absences, respectively. 
This factor is mainly added to mitigate the data imbalance between the two classes (which roughly occurs within our dataset in $10\%$ and $90\%$ of cases, respectively).  

Note that the eGPD log-density is multiplied by the binary variable $\ell_i$ because it is only defined whenever a landslide occurs (i.e., when $\ell_i=1$), which also naturally respects the obvious dependency between landslide occurrences and sizes. 
When $\ell_i=0$, the rightmost expression in \eqref{eq:loss} cancels out and need not be computed. 

\subsubsection{Model Training}
To efficiently train the model with backpropagation, we used the popular Adam optimiser \cite{kingma2014adam} with an initial learning rate of $10^{-3}$, which was further decayed every $50000$ step by a factor of $0.95$ until the model converged.
Various training batch sizes were tested, but with a batch size of $N=2048$ the model converged quickly and prevented the vanishing gradient problem.
The entirety of the dataset was randomly divided into training and test sets, which consisted of $70\%$ and $30\%$ of the total data, respectively.
Furthermore, the training set was randomly divided into a validation set of $30\%$ at every epoch to test the models' generalisation capability.
Further details on the sets of hyperparameters used to train the model are reported in the code-base. 

\subsubsection{Evaluation Metrics}
To evaluate the model performance, we considered two approaches. Firstly, we used the area under the curve (AUC) of the receiver operating characteristic curve \cite<ROC;>{hosmer2000} to assess the model's classification results, i.e, the probability $p(s,t)$ corresponding to the presence/absence of landslides in a given SU $s$ and year $t$. 
This is a well-known and accepted method in landslide susceptibility modelling \cite{chen2018landslide,dahal2023explainable}. 
Second, to validate the intensity component $i_q(s,t)$, we followed the approach presented by \citeA{cisneros2023deep}. 
The authors use quantile-quantile (Q-Q) plots to visually assess whether quantiles from the estimated eGPD distribution of the wildfire magnitude and qualitatively match their empirical counterparts.
In this paper, we did the same but used Q-Q plots for the areal density instead. 
To get a more precise and quantitative model diagnostic, we also used the continuous ranked probability score (CRPS), which is a proper scoring rule widely used to assess the calibration and sharpness of probabilistic forecasts \cite{gneiting2007strictly}. 
Suppose that, at a given space-time point $(s_i,t_i)$, our deep-learning regression model provides an estimated probability distribution $\hat F_i$ for the size variable $A(s_i,t_i)$, and denote the observed area density at $(s_i,t_i)$ as $a_i$. Then, the CRPS is defined as
\begin{equation}
\label{eq:crps}
{\rm CRPS}(\hat F_i, a_i)=\int_{0}^{\infty}\{\hat F_i(x)-\mathbb I(a_i\leq x)\}^2 {\rm d}x,
\end{equation}
where $\mathbb I(\cdot)$ is the indicator function taking value $1$ if the argument is true and $0$ otherwise. Note that the CRPS is minimised (i.e., it equals zero) when $\hat F_i$ is the degenerate distribution that puts all its mass at $a_i$; therefore, low values of the CRPS indicate that the estimated distribution $\hat F_i$ is not only well-calibrated with the observed value $a_i$, but it also has low variance. To assess the overall model fit over space and time, we then sum up all CRPS values for all observed data, i.e., we compute $\sum_{i=1}^n{\rm CRPS}(\hat F_i, a_i)$.

\section{Results}
This section reports the model's performance, its application for hazard modelling, and its use for future landslide hazard predictions in three separate sub-sections. 
\subsection{Model Evaluation}
\begin{figure}[t!]
	\includegraphics[scale=0.60]{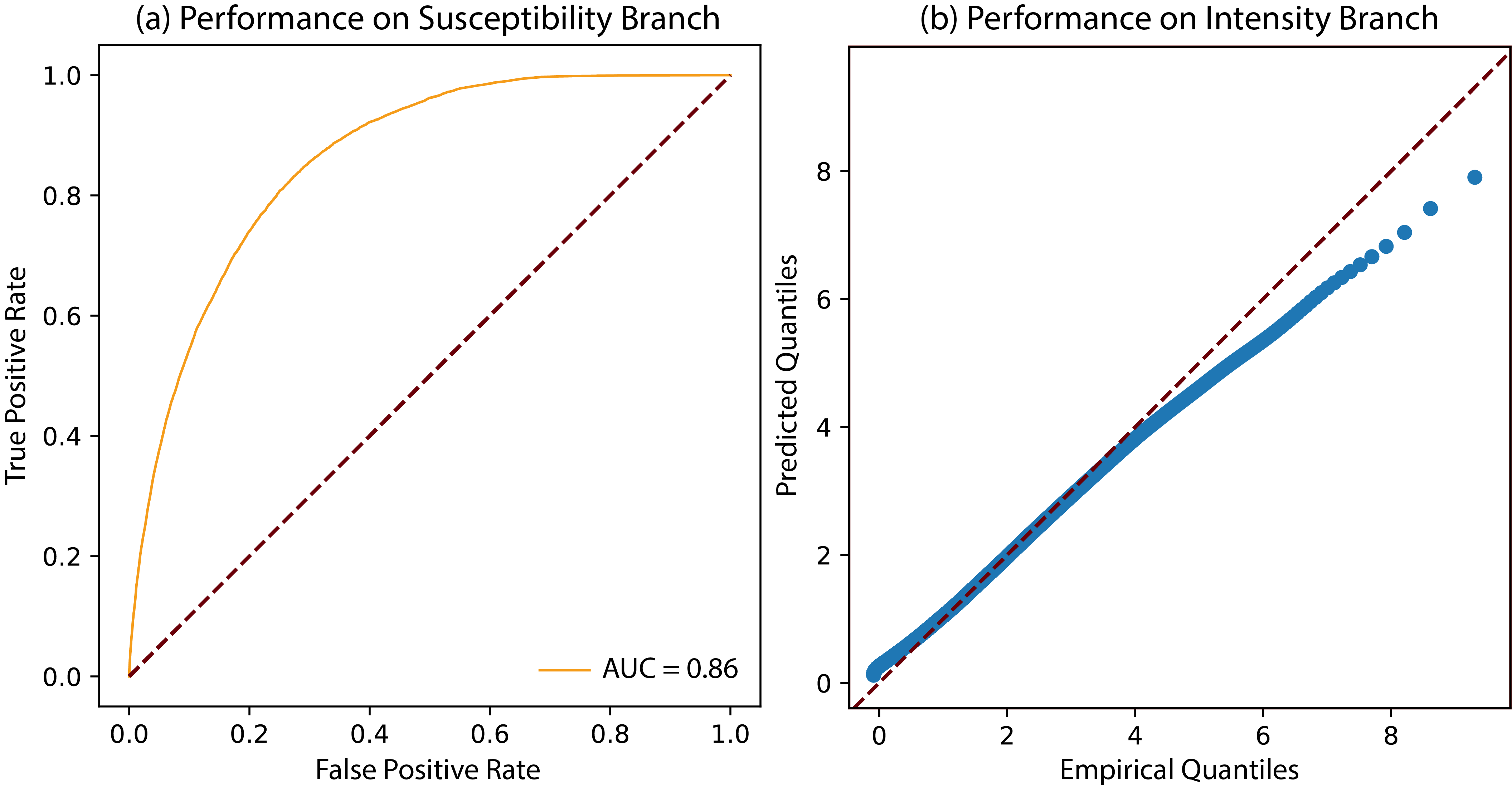}
	\caption{Overall performance of the developed model. (a) The model performance on the classification of landslides (i.e., landslide susceptibility), and (b) the Q-Q plot of displaying model-based vs empirical quantiles.}
	\label{fig:qqplot}
	\centering
\end{figure}

With the fitted model, we evaluated the model's performance by assessing the AUC score for the susceptibility ($p(s,t)$).
The intensity component ($i_q(s,t)$) was evaluated through CRPS scores and Q-Q plot. 
Figure~\ref{fig:qqplot} shows the ROC curve and the Q-Q plot. The model shows excellent predictive performance with an AUC value of 0.86 and a CRPS score of 3.25. We can observe that the Q-Q plot has an overall good fit with almost a 45$^\circ$ line, though there is some slight deviation in the upper quantiles, indicating that our model has a tendency to underestimate the size of the biggest landslides. Neverthess, the model provides a satisfactory performance overall.

\subsection{Current Hazard Estimates}
With the trained model, we predicted the (hypothesised) landslide hazard $h_{q;P}(s)$ over the study area for four different return periods $P$ and severity levels $q$. Specifically, we considered four return periods (with $P=5,10,15,20$ years).
These return periods are directly linked to the return period for the landslide hazard itself, given that precipitation is the primary trigger in this case. 
Moreover, we computed the landslide hazard for the three different severity levels, setting $q=5\%,50\%,95\%$ to understand different possible landslide hazard scenarios.
We recall here that each level of $q$ represents approximately 0.02\%, 0.42\%, and 4.25\% of the area density per SU, respectively.

\begin{figure}[t!]
	\includegraphics[width=1.0\textwidth]{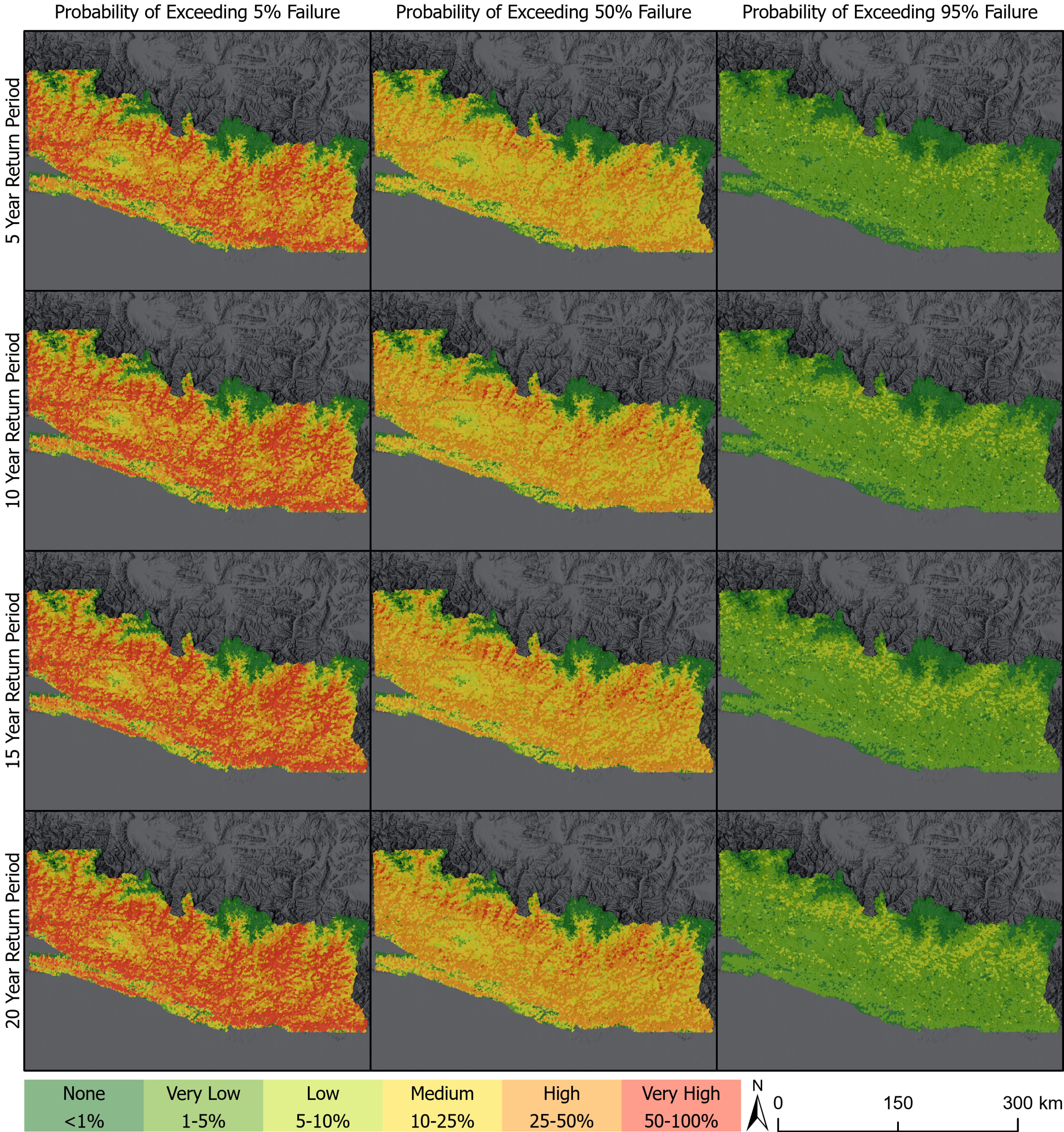}
	\caption{Predicted landslide hazards for four return periods and three exceedance thresholds. It represents how landslide hazard changes over higher intensity of precipitation and over different $q$ thresholds.}
	\label{fig:predcurrent}
	\centering
\end{figure}

The estimated hypothesised landslide hazard based on historical precipitation data (``current scenario'') is shown in Figure~\ref{fig:predcurrent}. 
In a nutshell, we can observe that the landslide hazard generally increases with higher return periods. 
When $q=5\%$, this increase is particularly visible because the estimated hazard $h_{q;P}(s)$ often changes from 10--25\% to 50--100\%.
Similarly, when $q=50\%$, for most of the area $h_{q;P}(s)$ changes from 5--10\% to 25--50\% and when $q=95\%$, the same changes from 1--5\% to 5--10\%. 
When $q=95\%$, we can observe that the majority of the area has extremely small hazards (category ``None'' and ``Very low'') for the 5-year return period while the majority of the locations have a ``Low'' and ``Very Low'' hazard for the 20-year return period. 
In other words, irrespective of the column one looks at, from the first row to the last, we can see an increase in the expected hazard (the colours shift towards the red side of the spectrum).
Looking at the plot across rows shows a shift towards the left side of the hazard colorbar, this being due to the fact that the probability of a failure to cover most of a SU becomes increasingly unlikely. 

\begin{figure}[t!]
	\includegraphics[width=1.0\textwidth]{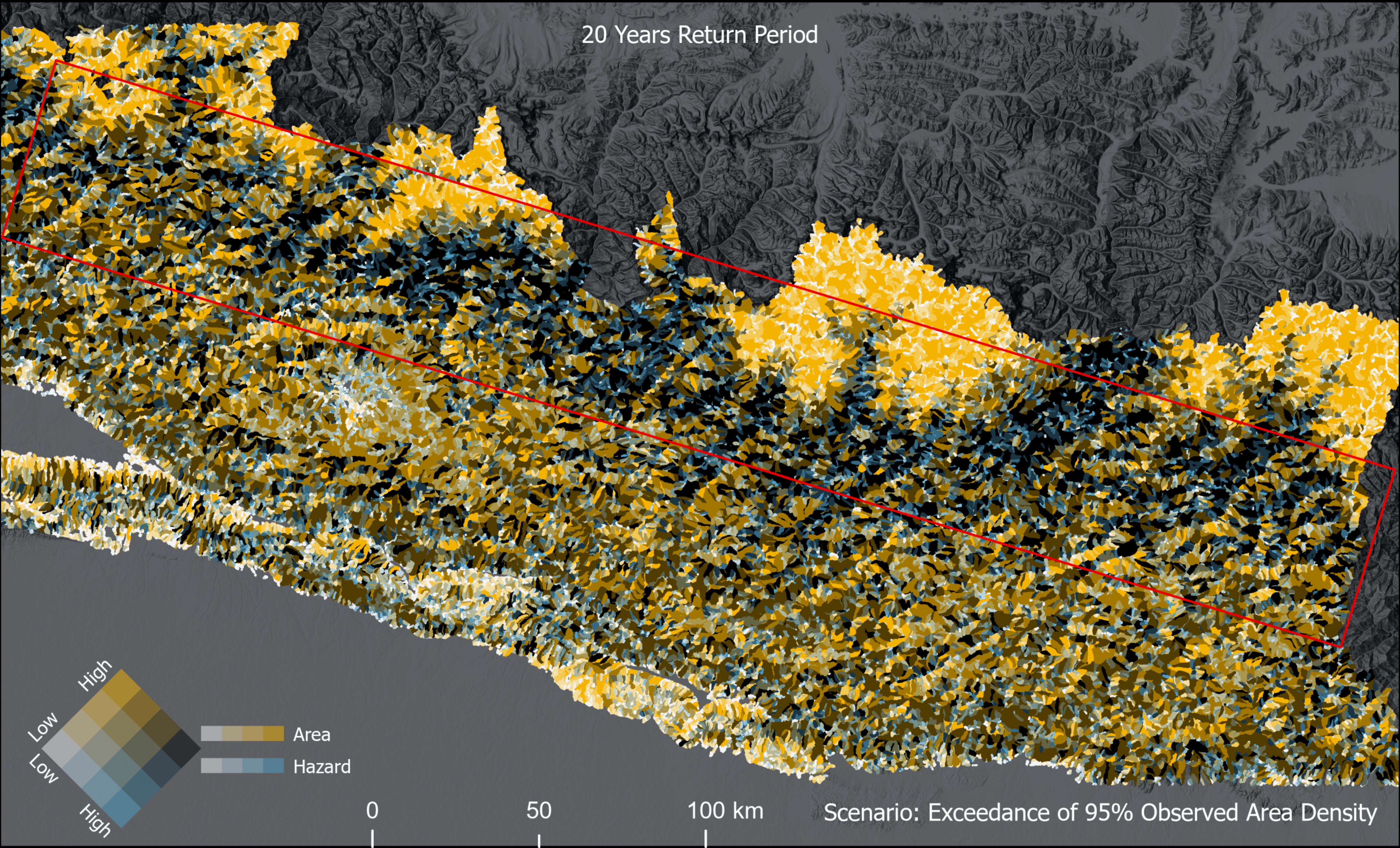}
	\caption{Landslide hazard with $q=95\%$ for a 20-year return period precipitation event, plotted jointly together with the SU area. The area delimited by the red lines shows the Himalayan region with a high landslide hazard associated with the largest SU.}
	\label{fig:usecase}
	\centering
\end{figure}

Figure~\ref{fig:predcurrent} alone, however, does not provide a complete overview of the expected landslide hazards because we do not know how big the SUs are, thus limiting the true extent of landslide hazard information. 
A possible solution to mitigate this visualization issue is presented in Figure~\ref{fig:usecase} for a 20-year return period hazard with a $q=95\%$ exceedance threshold, representing the worst case scenario from Figure~\ref{fig:predcurrent}.
This is intended to represent how we can use landslide hazard predictions in decision-making.
For instance, we can observe dark-coloured locations with high SU areas and high hazard, meaning those locations need particular attention in terms of potential mitigation measures.
In fact, if the SU area is large and the model estimates a large hazard for it, this implies that one should expect a much larger landslide compared to a large area density predicted for a small SU. 
Additionally, in Figure~\ref{fig:usecase}, a dark band (within the red polygon) appears in northern Nepal. This region is characterized with the highest hazard with the largest SU areas and is located in the upper-mountain region of Nepal, which is affected by landslides every year. 
This representation can be further used for landslide risk purposes by overlaying the information related to population density, buildings, economic assets, etc.
As a result, one could assess the exposure of elements-at-risk at varying levels of landslide hazards. 

\subsection{Future Hazard Projections}
\begin{figure}[t!]
	\includegraphics[width=1.0\textwidth]{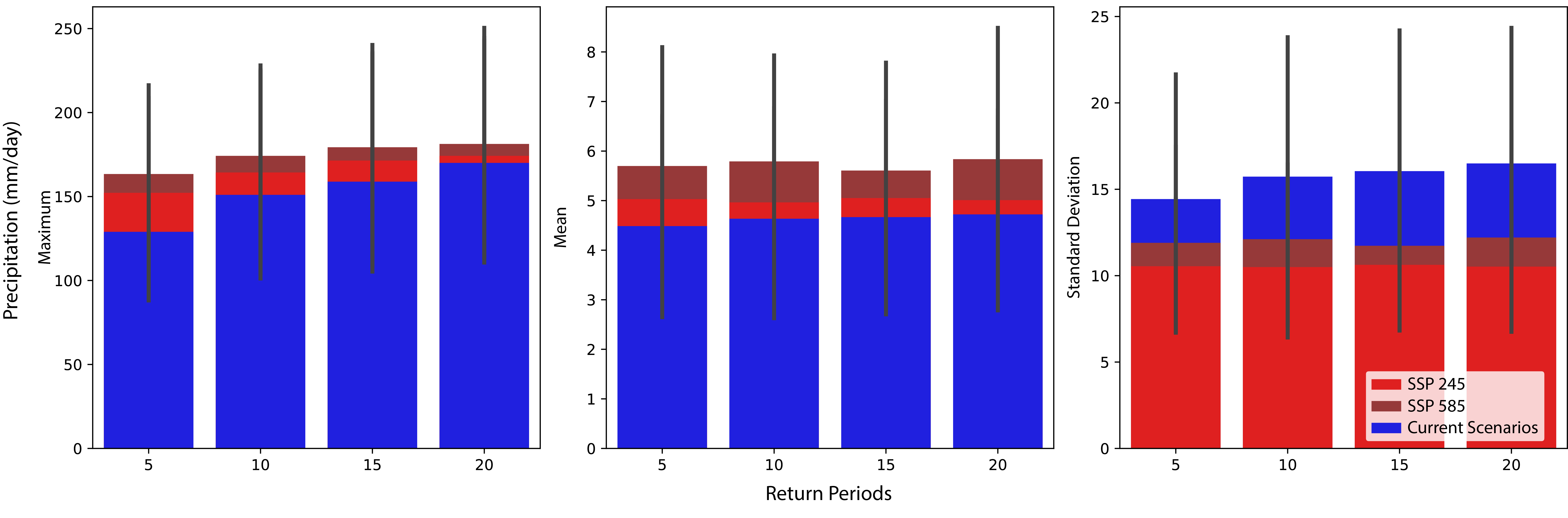}
	\caption{Return levels of the annual maximum (left), mean (middle) and standard deviation (right) of overall precipitation in the study area in the current scenario (historical period, blue) as well as future projections under different climate change pathways (SSP245, red, and SSP585, dark red) for four different return periods ($P=5,10,15,20$ years, from left to right bars). The vertical dark lines represent the range of precipitation values.}
	\label{fig:Fig8}
	\centering
\end{figure}

Before showing the estimated landslide hazard under various climate change scenarios, we explore the change in the rainfall return levels for future precipitation (under both SSP245 and SSP585 pathways) compared to the current situation. Figure~\ref{fig:Fig8} reports the results. 
We can observe that the mean precipitation has particularly increased in all climate change scenarios compared to the current situation (historical period). 
For the annual maximum precipitation, all return periods have major increases except for the 20-year return period. 
Moreover, for the SSP245 and SSP585 pathways, the difference in maximum and standard deviation is small, with SSP585 having slightly higher values in all cases, whereas the SSP585 has a major increase in mean precipitation compared to both the current scenario as well as the SSP245 pathway.
Since this research focuses on landslide hazard modelling and the climate change scenarios are mostly used to test our model, we do not focus much on the reasons behind such climatic behaviour, and readers should recall that downscaled climate projection data are usually affected by bias and uncertainties which may lead to hazard misrepresentation. 

\begin{figure}[t!]
	\includegraphics[width=1.0\textwidth]{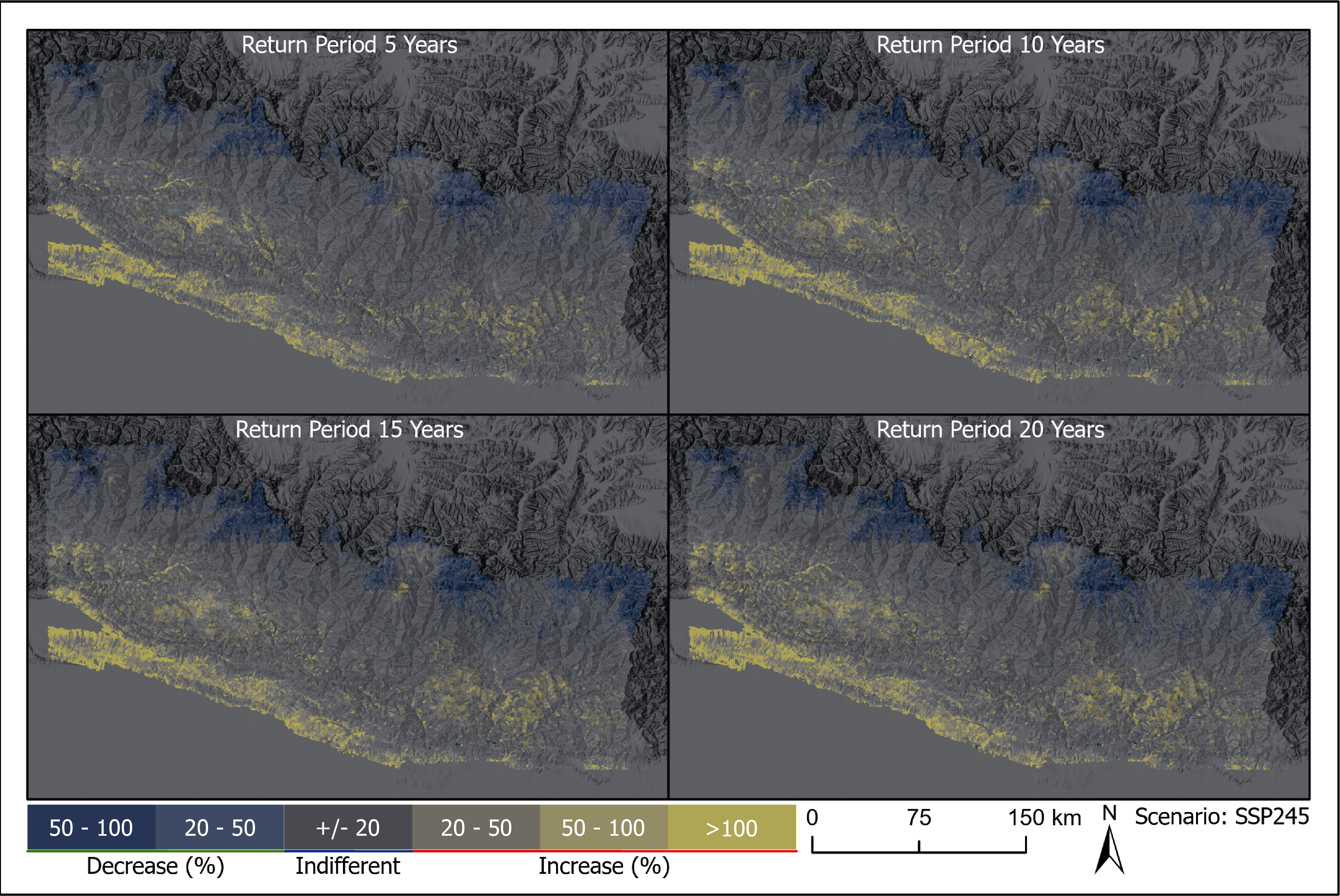}
	\caption{The change in landslide hazard for the $q=50\%$ exceedance threshold and different rainfall return periods (different panels) under the SSP245 climate change pathway.}
	\label{fig:Fig9}
	\centering
\end{figure}

\begin{figure}[t!]
	\includegraphics[width=1.0\textwidth]{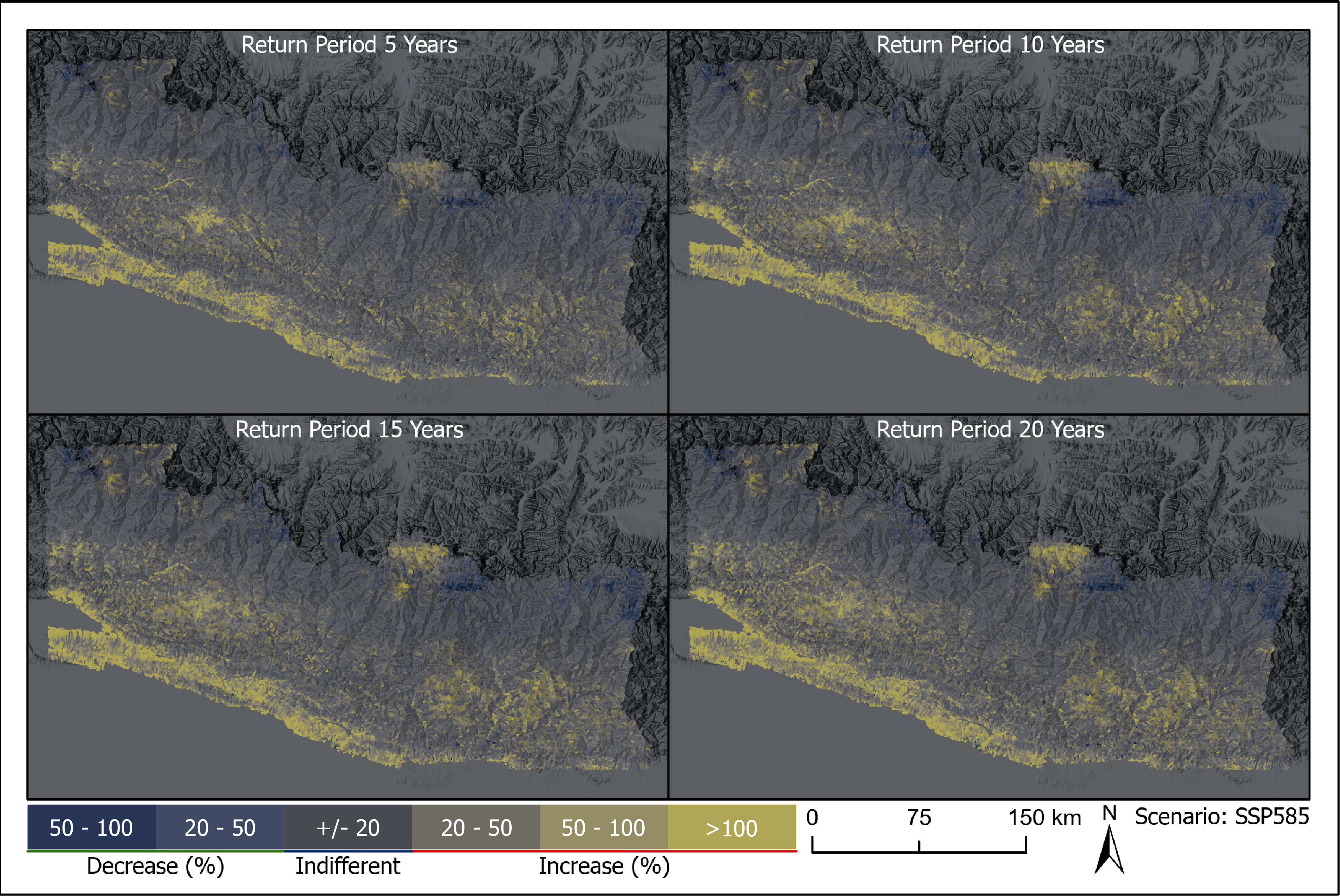}
	\caption{The change in landslide hazard for the $q=50\%$ exceedance threshold and different rainfall return periods (different panels) under the SSP585 climate change pathway.}
	\label{fig:Fig10}
	\centering
\end{figure}

With the designed precipitation for climate change scenarios, we estimated the landslide hazard for the next 80 years (from 2020) for four different return periods.
Since it is more meaningful to visualise how the landslide hazard changes over different climate change scenarios than to see the actual hazard values, we plot the change in the median landslide hazard (i.e., with $q=50\%$), compared to current scenarios in Figure~\ref{fig:Fig9} (SSP245) and Figure~\ref{fig:Fig10} (SSP585). 
Interestingly, most locations under both climate change scenarios experience a change within the $\pm20\%$ range. 
We classify this range as a ``no-change'' range because our dataset has many uncertainties that do not allow us to monitor relatively small changes in the hazard. 
These include the uncertainties in climate projections and their bias with respect to the CHIRPS dataset, as well as the uncertainties and bias in CHIRPS data itself.
Nevertheless, under both SSP245 and SSP585 scenarios, the landslide hazard increases in the middle and lower Himalayas (represented by the lower half of the study area). 
We can also observe that under both climate change scenarios, the hazard increases with larger rainfall return periods. 
Interestingly, the hazard decreases in the upper Himalayas under the SSP245 pathway, while this is less prominent under the SSP585 pathway. 
In the SSP585 case, we further see that in the upper Himalayas, the hazard increases, which is not the case in SSP245 cases. 
We can further observe that the spatial pattern of changing landslide hazards is similar in both pathways, even though their spatial density is different. 

\section{Discussion}

\subsection{Interpretation}
Our model is able to satisfy the landslide hazard definition, testing it over 30 years of landslide history in Nepal. 
The advantage of generating exceedance threshold hazard estimates at various return times mainly translates into obtaining a full picture of the landslide process in a complex terrain such as the Himalayas. 
Figure \ref{fig:predcurrent} decomposes the probability of the hazard exceeding a given landslide density per SU (looking at single rows). 
This highlights a generic trend where most of the Nepali landscape appears to be potentially subjected to high probabilities of landslide occurrence, associated with high expected landslide area densities. 
This trend decreases moving from the left to the right panels. 
The last panel boils down to the worst-case scenario, the typical target of hazard and risk models, where a few large hazard hotspots are marked in the highest sector of the Himalayan belt. 
These translate into the probability of a failure to mobilize more than the 95$^{th}$ percentile of the landslide area density based on the past 30 years' observations, which naturally is a very unlikely phenomenon.  
Exploring the same figure vertically, one can see variations in landslide hazard estimates at increasing return times. 
Notably, the 5-year return time consistently represents the best-case scenario, which increasingly worsens moving towards the 20-year return time map.    
An important visualization tool is offered in Figure \ref{fig:usecase}, where the estimated hazard for the current scenario is combined together with the SU area, in a bivariate colorcoded map. 
What stands out the most is that the highest section of the upper Himalayas (displayed in yellow) present a relatively low landslide hazard, associated with relatively small SUs. 
This is mainly due to the fact that these areas are permanently frozen when it comes to interpreting the low hazard estimates. As for the small SU extent, this is due to the heavily tectonized and thus dissected landscape, which in turn produces small slopes.
Therefore, the hazard at these locations appears to be reasonably low overall. 
Looking at the section between upper and middle Himalayas (encompassed in the red polygon), one can notice the darkest among the bivariate colorcodes. 
This implies that not only is the hazard estimated to be high there, but also the SU size is at its highest. 
Because we model landslide hazard as a function of area density per SU, a higher percentage of SU failing, associated with a large SU, has the potential to release large and very dangerous mass movements.
As for the topographic section that occupies the middle and low Himalayas, this is associated with a mixture of hazard and SU size estimates.
Therefore, here it is more difficult to cluster a single behavior and the spatial pattern of the two parameters becomes more erratic. 
To conclude the interpretation of our model output, we focus on Figures \ref{fig:Fig9} and \ref{fig:Fig10}. 
There, we geographically translated the output of our model when used for simulation purposes. 
Specifically, plugging-in the spatio-temporal rainfall signal coming from climate projections, we obtained scenarios of landslide hazard for two cases where: \textit{i}) the underlying model assumes that climate protection measures are being taken to mitigate global warming; \textit{ii}) the underlying model assumes our society continues to develop, relying on fossil fuels.
What stands out is that the conservative climate pathway (SSP245) depicts an increase with respect to the current scenario along the lowest portion of the Himalayan belt. 
Moreover, a marked decrease along the top of the topographic profile becomes evident (areas in blue) across all return times under consideration. 
When comparing this projection to the most environmentally negative pathway, one can notice an even larger increase in the landslide hazard across all respective return times. 
Interestingly, the upper Himalayas also experience an increase, with the blue areas diminishing significantly.

\subsection{State of the art}
In the case of landslide hazard modelling, the seminal work of \citeA{guzzetti2005probabilistic} 
used a probabilistic approach to separately solve three different components of the landslide hazard equation \cite{guzzetti1999landslidehazard}.
Their approach predicted susceptibility over a SU partition and estimated the area density and frequency through an empirical distribution function rather than conditioning it to environmental factors. 
The major limitation of that work was to have a fixed distribution of landslide area density, which might not hold as the intensity of triggering events changes over time due to many factors, such as climate change.
Moreover, using a fixed empirical distribution for both intensity and frequency did not reflect the spatio-temporal characteristics of the data. 
In fact, assuming that all slopes are capable of having the same frequency and intensity of landslide is a strong assumption that may hold only theoretically rather than in a realistic experiment. 

Improving on this, our model attempted to predict the landslide intensity as a function of environmental conditions, while jointly considering the dependence between landslide occurrence and size. 
As for the frequency component, we assume the frequency of the triggering event instead of modelling the frequency of landslides directly, which explicitly recognises landslides as a cascading event.

This approach, in particular, is a more flexible solution with potential applications to modelling the future evolution of landslides. 
Compared to other works that focus on intensity as a landslide area density and counts \cite{guzzetti2002power,lombardo2021landslide,aguilera2022prediction,bryce2022unified,yadav2022joint}, this approach provides a complete picture of the landslide hazard by modeling their occurrence probability as well as intensity exceedance probability both over space and time.

Moving beyond the comparison with respect to the work of \citeA{guzzetti1999landslidehazard}, below we list improvements with respect to other research contributions. 
Let us recall that classical susceptibility models only estimate the probability of landslide occurrence locations, under the assumption that such value does not change in time.
Our proposed model introduces elements of spatio-temporal analyses, together with a joint susceptibility-intensity estimation.  

With respect to examples of intensity models built for large regions, most of them also perform purely spatial operations, whether the intensity is expressed as counts \cite{Lombardo2019} or landslide area \cite{lombardo2021landslide} per SU. 
Even in this case, we believe our space-time extension does provide a more informative hazard estimation. 
Few examples do exist where the probability of landslide occurrence and area density is addressed. 
This is the case of \citeA{fang2024landslide}.
However, the authors built two separate models, one for the susceptibility component and one for the intensity. 
Our model expands on their research by jointly modelling susceptibility and intensity and also because we fitted an eGPD rather than a log-Gaussian model for the area density component. 

\subsection{Supporting Arguments}

Moreover, our approach allows one to monitor the full spectrum of landslide hazards for different trigger and exceedance thresholds. This is not only limited to the observed realm but also to the time and intensity beyond the range of the observed data, because it is well supported by the extreme value theory, which is designed to do so \cite{naveau2016modeling}. 
From the planning perspective, this may well be the most interesting element of our proposed model \cite{ozturk2019network}. 
Any engineering design requires, by definition, a specific return time to use as a mark for the construction to be deemed reliable and usable. 
For instance, bridges are designed knowing the estimates of flood return times, which are usually translated in water heights, which the bridge needs to be comparatively higher. 
In a similar manner, landslides share the same requirements. 
For instance, \citeA{popescu2009engineering} recommend designing retaining walls with a 100--year return period when subjected to the stress of a moving landslide body.
However, currently the calculation of exceedance probability related to landslides has been mostly confined to site-specific conditions. 
Our contribution allows one to extend this paradigm towards large geographic extents. 
What should be stressed at this stage is the fact that for engineering design, the landslide intensity is typically assessed in terms of landslide velocity, kinetic energy or impact force. 
These are physical quantities that can be naturally brought into geotechnical calculations, as they share the same dimensions the construction is based on. 
However, estimates on landslide density are still one step away from being translatable into useful information for engineering design. 
Future development could involve expressing the landslide intensity not in the form of area density but rather as velocity. 
However, such an experiment should involve first running spatially and temporally distributed simulations of landslide kinematic behaviors, from which an architecture such as the one presented in this contribution could learn and further simulate hazard.

Aside from the future engineering applicability of our model, we see a more straightforward use for master planning. 
Specifically, our model could be used to prioritize land use development plans because decision-makers will have the full spectrum of landslide hazard and its expected temporal variation at their disposal. 
Therefore, they can decide whether to invest in specific projects minimizing the risk their capital will be subjected to. 

Taking a step back and focusing on computational aspects, non-linear methods such as neural networks
typically require substantial resources. 
However, our model can be scaled up to take advantage of high-performance-computing solutions that involve GPU-heavy machines. 
In such case, the number of input features and data points can be increased to the point of supporting regional or even global scales (assuming landslide polygonal inventories with good quality and completeness are available). 

For example, inference in the entirety of our study area takes $\approx$ 7.25 seconds in a Pascal P100 GPU, which is achieved via a parallelised and GPU-compatible model and data pipeline. Thus, this model ensures scalability to larger spatiotemporal regions without a significant performance plunge.

\subsection{Opposing arguments}

Despite the strengths we highlighted in the previous sections, our modeling approach still has some limitations which are worth examining and reflect on as the starting point of future development. 

First, we have based our model on the eGPD distribution. 
This has performed quite well in the case of the spatio-temporal data we examined, but it is not guaranteed to be
the case elsewhere. 
Therefore, further tests in other geographic contexts and with other landslide types should be carried out to examine if the eGPD assumption still holds or whether other distributions inspired by statistics of extremes should be implemented as well. 
It is worth stressing that the way we have scripted our modeling protocol can be extended to other parametric distributions (e.g. Gumbel, Weibull) with relative ease. 

Aside from the technical aspects, something else we should highlight involves the scenario-making procedure we put in place. 
Currently, we have assumed that the return period of the landslide hazard is the same as the return period of the triggering rainfall extremes. 
This is possible because our inventory only considers the landslides caused by rainfall, not other anthropogenic and seismic factors. 
This is an exceptional case, and it is not common to inventories in other regions. 
Thus, when there are multiple triggering factors, it is ambiguous which triggering event and respective return period to use. 
Moreover, how to combine multiple return periods of different triggering events in a cascading hazard is an ongoing research question for which we do not know the answer yet. 
For example, to understand existing hazards, the combined effect of predisposing factors such as road construction, precipitation, and ground shaking can be included in the model as input covariates, but when we need to include them in future scenarios, identification of their combined effect and return period is extremely challenging. 
As a result, using our model as is may require some simplification in the form of limiting the return period of landslides to the return period of the primary triggers, such as precipitation or earthquakes.

\section{Conclusion}
\label{sec:conclusion}

Our statistical deep-learning model formally satisfies the landslide hazard definition proposed by \citeA{guzzetti1999landslidehazard} for the first time in a single architecture. 
This is achieved via a joint model capable of explaining the spatio-temporal variability of the landslide occurrence distribution as well as the associated areal density per SU. 
Solving for the susceptibility component has become a standard in the literature, while addressing the intensity element is much less consolidated. 
Here we solve for the intensity with an original modeling archetype that flexibly estimates the two tails and the bulk of the distribution via the eGPD model. 
Another element of innovation is in the estimation of the eGPD, whose parameters varying through a complex interaction of spatio-temporal covariates are retrieved via Neural Networks.
Ultimately, we test our model as a simulator for scenario-based hazard assessment, plugging in climate projections at specific return times.   

It is important to stress that in space-time modeling, one can contract or expand time and space depending on the data, modeling target and research question at hand. 
Currently, we tested our model partitioning space via SUs and partitioning times on a yearly basis. 
A potential extension to be tested would require shrinking time on a daily basis (assuming daily landslide polygonal inventories are available). 
In such a case, rather than modeling long-term landslides, one could generate predictions on a more frequent basis, de facto converting our model into an early-warning-system equipped with a full landslide hazard estimation. 

Ultimately, we recall that the most recent definition of landslide risk \cite{corominas2023revisiting} requires the recurrence interval to quantify landslide hazard.
Our model solves for the hazard definition allowing one to simulate for specific return times. 
Therefore, differently from other data-driven applications, our model can be directly of use for risk assessment protocols. 
This further adds to the positive aspects listed above, although its inclusion in risk assessment is yet to be verified. 

\newpage

\section{Open Research}
The data and code used in this research are available via the FAIR repository at https://doi.org/10.5281/zenodo.10567233 and https://zenodo.org/doi/10.5281/zenodo.10567256, respectively.

\acknowledgments
We would also like to thank the KAUST competitive research grant (CRG) office for funding support for this research under grant URF/1/4338-01-01. This work used the Dutch national e-infrastructure with the support of the
SURF Cooperative using grant no. EINF-7984


%
%



\bibliography{landslides.bib}

\end{document}